\documentclass[letterpaper, 10 pt, conference]{ieeeconf}  

\IEEEoverridecommandlockouts                              

\usepackage{graphics} 
\usepackage{epsfig} 
\usepackage{mathptmx} 
\usepackage{times} 
\usepackage{amsmath} 
\usepackage{amssymb}  

\usepackage{subfigure}
\usepackage{multirow}
\usepackage{kotex}
\usepackage{mathtools}
\usepackage{booktabs}
\usepackage{wrapfig}
\usepackage{titlesec}

\usepackage{capt-of}
\usepackage{cuted}

\usepackage{graphicx}
\usepackage[dvipsnames]{xcolor}

\title{\LARGE \bf
TERDNet: Transformer Encoder-Recurrent Decoder Network for Scene Change Detection
}

\author{Jiae Yoon and Ue-Hwan Kim$^{*}$%
\thanks{All authors are with the Department of AI Convergence, Gwangju Institute of Science and Technology (GIST), Gwangju 61005, Republic of Korea.
        {\tt\small jiaeyoon@gm.gist.ac.kr}}%
\thanks{$^{*}$Corresponding author: Ue-Hwan Kim ({\tt\small uehwan@gist.ac.kr}).}%
}

\begin{document}
\thispagestyle{empty}
\pagestyle{empty}

\maketitle

\begin{strip}
\centering
\vspace{-1.7cm} 
\includegraphics[width=\textwidth]{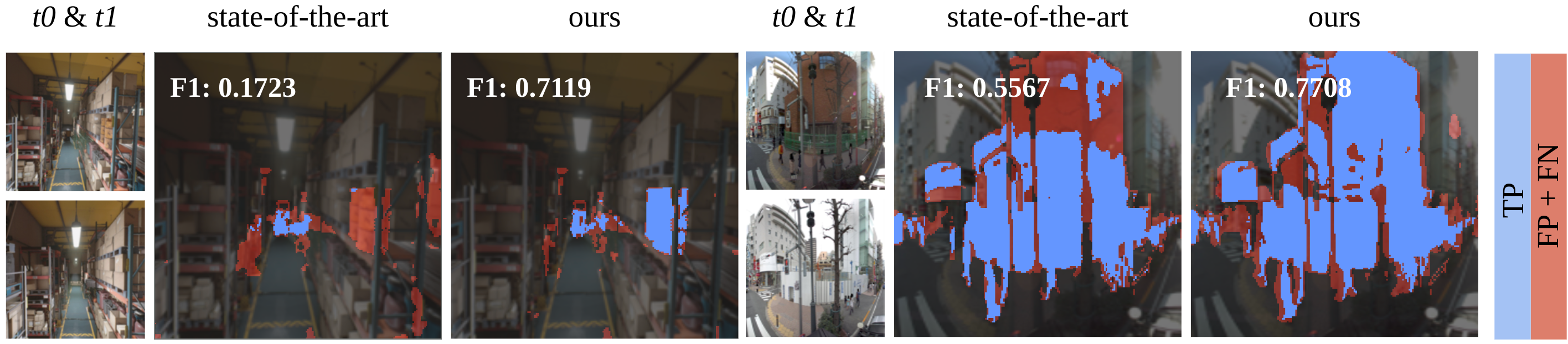}
\captionof{figure}{
\textbf{Comparative results of the current state-of-the-art C-3PO \cite{wang2023reduce} and our TERDNet on four benchmark datasets.} 
TERDNet achieves superior quantitative performance and produces more precise change masks with clearer boundaries compared to the existing state-of-the-art approach.
}
\label{fig:teaser}
\end{strip}

\begin{abstract}
In this work, we address the challenge of Scene Change Detection (SCD), where the goal is to identify variations between two images of the same location captured at different times. Existing SCD models often overlook the varying importance of features across layers, employ single-step decoders that confine refinement, and provide limited insight into encoder pretraining strategies. We propose TERDNet, a Transformer Encoder–Recurrent Decoder Network designed to overcome these limitations. TERDNet consists of a transformer-based encoder that extracts multi-level representations, a feature fusion module that integrates correlation volumes with these features, a recurrent 3-gate-GRU decoder that performs iterative refinement, and a combined convolution–interpolation upsampler that restores fine-grained resolution. Extensive experiments on four public benchmarks show that TERDNet consistently outperforms prior approaches and produces more accurate and detailed change masks. Ablation studies confirm the benefit of segmentation-based pretraining and the effectiveness of our fusion design. In addition, robustness tests under viewpoint misalignment confirm TERDNet’s potential for deployment in real-world robotic systems, where reliable perception is critical. Our code is at {\tt\small https://github.com/AutoCompSysLab/TERDNet}.

\end{abstract}


\section{Introduction}
Scene Change Detection (SCD) generates a change mask indicating areas of variation from two images captured at different times but depicting the same location \cite{radke2005image}. In robotics, SCD is particularly valuable as it enables mobile agents to reason about dynamic environments \cite{yew2021city}, identify object-level changes in real-world deployments, and adapt navigation strategies accordingly \cite{kannan2025zeroscd}. Further, autonomous robots and vehicles operating in urban or indoor environments must continuously assess environmental shifts \cite{rowell2024lista}, such as moved furniture or newly placed obstacles, to make informed decisions and ensure safe traversal \cite{looper2023vsg}. SCD is similar to the segmentation task \cite{kirillov2023segment} in that it discriminates the class of each pixel. The two tasks share the commonality of requiring sophisticated edge extraction, but they also differ in that SCD involves inputting two images; in comparing two images with a time difference, SCD resembles optical flow \cite{wu2024lightweight}. As a result, most SCD models adopt an encoder-decoder architecture similar to segmentation and optical flow models.



Conventional SCD studies leveraging deep neural networks \cite{wang2023reduce,gan2024rfl} have predominantly employed multi-level feature maps derived through feature pyramids \cite{lin2017feature}. This approach does not explicitly account for the varying importance of the feature maps extracted from each layer.
With the emergence of transformer-based models, recent SCD works \cite{cho2025zero, lin2025robust, kannan2025zeroscd, kim2025towards} have begun adopting transformer encoders. However, prior studies have not systematically analyzed which pretraining strategies are most suitable for SCD. On the decoder side, the prevailing architecture within SCD models typically employs single-update decoders. While this approach is computationally efficient, it can limit the models' ability to generate detailed and accurate change masks. Single-update decoders are constrained in their capacity to refine and enhance the output through iterative processes, which can significantly detract from the overall effectiveness and accuracy of SCD models.

To overcome these limitations of SCD models, we propose the Transformer Encoder and Recurrent Decoder Network (TERDNet). TERDNet incorporates a feature fusion module that integrates correlation volumes with multi-level transformer features. This module also learns the relative importance of each layer’s feature map, leading to better fusion of features across layers. In addition, we conduct experiments on the encoder, including comparisons between CNN and transformer backbones as well as different pretraining strategies, to analyze their impact on SCD.

Further, we devise a recurrent structure for the decoder to perform iterative refinement, which leads to performance enhancement. While prior optical flow models \cite{teed2020raft, zhou2023mvflow} have demonstrated the benefit of convGRU for iterative refinement, these approaches have not been designed for SCD. In contrast, our proposed 3-gate-GRU explicitly incorporates the feature pyramid and dynamically weights layer importance, enabling recurrent refinement tailored specifically to SCD. The 3-gate-GRU receives inputs from the current input, past output, and the feature pyramid with applied weights. As illustrated in Fig. \ref{fig:teaser}, this design addresses the limitations of existing SCD decoders by enabling iterative updates that refine the output progressively.

In summary, the main contributions of our work are as follows:
\begin{itemize}
    \item \textbf{Encoder and Feature Fusion Module}: We leverage transformer encoders to extract multi-level features and introduce a fusion module that integrates correlation volumes with these features, while explicitly learning the relative importance of each layer’s representation.
    \item \textbf{Recurrent Structure in Decoder}: We propose a 3-gate-GRU tailored for SCD, which enables iterative refinement and dynamic integration of features from the transformer backbone.
    \item \textbf{SoTA Performance}: Our model achieves state-of-the-art results on multiple public benchmarks.
\end{itemize}

\section{Related Work}

\textbf{Change Detection.} Change Detection (CD) aims to identify differences in the state of an object or phenomenon by comparing data collected at different times \cite{sun2021nonlocal}. With the advent of deep learning, significant advancements have been made by leveraging the capabilities of CNNs and RNNs. For example, ChangeNet \cite{daudt2018urban} has employed a fully convolutional network to learn and detect changes directly from raw image pairs. ConvLSTM \cite{shi2015convolutional} has captured temporal dependencies, especially beneficial in video surveillance. L-UNet \cite{Sun2020LUNetAL} has enhanced CD performance by applying an Atrous convolution structure to convLSTM, thereby leveraging both temporal and spatial information. Huang et al. \cite{huang2023background} have enhanced weakly supervised learning by mixing background information to create more robust training samples.

\textbf{Scene Change Detection.} Scene Change Detection (SCD) is a task that identifies areas of change at the pixel level in images captured at different times, $t0$ and $t1$, of the same location. While CD encompasses various domains such as remote sensing \cite{gong2015change}, SCD focuses specifically on changes in visual scenes. 
In contrast to CD, where images often align well due to controlled or satellite-based capture, SCD must additionally contend with imperfect alignment, variations in lighting, and occlusions. These factors complicate direct pixel-level comparisons and make the task inherently more challenging.

Several works have led to the advancement of SCD. FC-Siam \cite{daudt2018fully} has applied a siamese network to SCD to process each image pair independently; the model, composed of a fully convolutional network (FCN) \cite{long2015fully}, has combined high-resolution features extracted from early layers with abstract features from deeper layers, inspired by U-Net \cite{ronneberger2015u}. 
CSCDNet\cite{sakurada2020weakly} has proposed that utilizes CNN encoder and correlation layers. DR-TANet \cite{chen2021dr} has introduced a dynamic receptive temporal attention module inspired by self-attention to represent the relationship between two feature maps. C-3PO \cite{wang2023reduce} has replaced the decoder of the SCD model with a segmentation model decoder from DeepLab \cite{chen2017deeplab} or FCN \cite{long2015fully}.
Conventional SCD models have predominantly employed CNN-based architectures. These architectures generally provide weaker representational capability than transformers. They also generate feature maps of varying resolutions, which forces uniform input into the decoder regardless of layer importance.

Recently, several SCD models have adopted transformer-based architectures. RobustSCD \cite{lin2025robust} utilizes the robust feature extraction capabilities of DINOv2 \cite{oquab2023dinov2}. Meanwhile, ZSSCD \cite{cho2025zero}, ZeroSCD \cite{kannan2025zeroscd}, and GeSCD \cite{kim2025towards} leverage the Segment Anything Model (SAM) \cite{kirillov2023segment}, a segmentation foundation model, to enable zero-shot SCD.
Although models leveraging foundation models can exploit richer representational power, prior studies do not provide an analysis of which pretrained weights of such models are most appropriate for SCD.
Furthermore, the simplicity of existing SCD decoders reduces computational costs but at the same time limits their ability to produce detailed change masks. This limitation indicates a significant opportunity for further advancements, particularly in the design of foundation model–based encoders and recurrent decoder structures.

\begin{figure*}
\begin{center}
\includegraphics[width=1\linewidth]{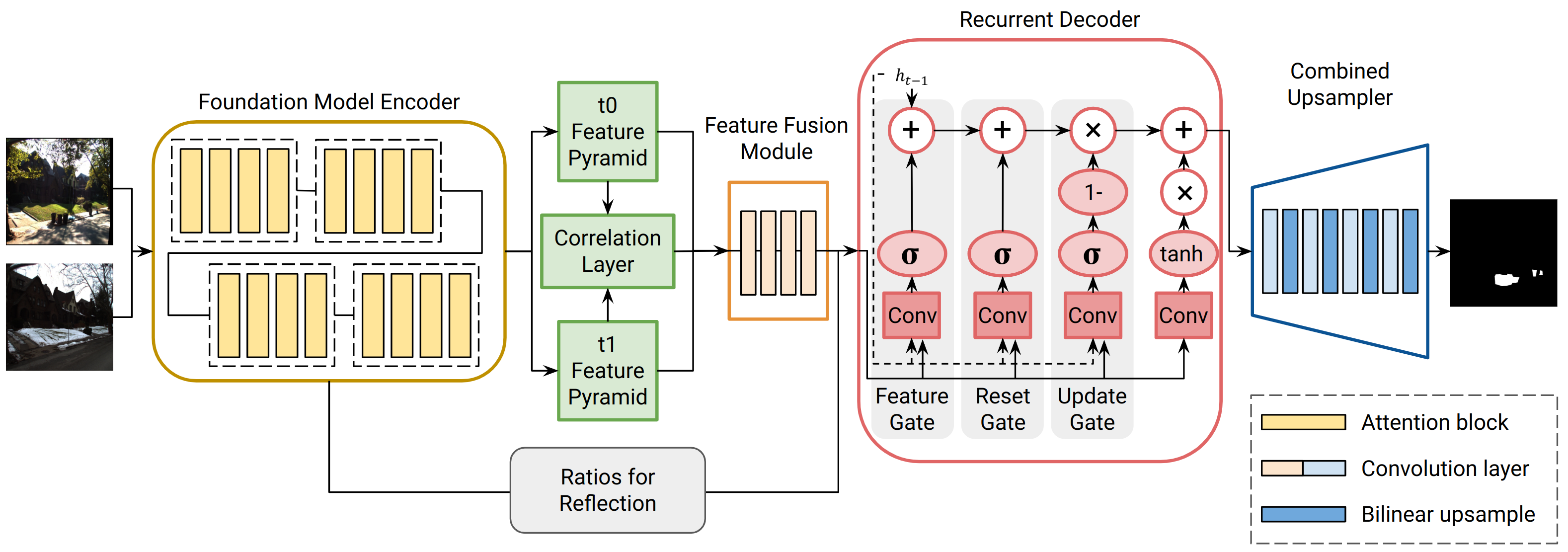}
\end{center}
\caption{\textbf{The architecture of the proposed TERDNet.} The foundation model-based backbone encoder extracts the feature pyramid from the two images. 
The decoder performs recurrent updates with the proposed GRU, and Ratios for Reflection computes the gating map $f_t$ in Eq.~(2) from pyramid differences.
The Feature Fusion Module combines the two feature maps and the correlation volume, feeding the combined feature into the decoder. The decoder uses three gates of the proposed GRU to determine the contribution weight of each source of information and performs recurrent updates. The combined upsampler alternates between convolution layers and bilinear interpolation and restores the output change mask to the original resolution.
}
\label{fig:main_fig}
\end{figure*}

\textbf{Foundation Models.} 
Foundation models refer to models trained on large datasets that can be applied to a variety of application tasks, with transformer-based foundation models recently gaining prominence \cite{kenton2019bert}. Following the introduction of transformer models based on attention modules in natural language processing (NLP) \cite{vaswani2017attention}, vision transformer models have emerged for vision tasks \cite{dosovitskiy2020image}. These models offer higher representation power than CNNs but demand large datasets due to their weaker inductive bias \cite{mormille2023introducing}. Vision foundation models utilizing the transformer structure have achieved high performance in various vision tasks such as classification \cite{liu2021swin}, object detection \cite{carion2020end}, and segmentation \cite{kirillov2023segment}, and researchers actively have utilized these foundation models.

\textbf{Recurrent Decoders.} 
Applying Recurrent Neural Networks (RNNs) to decoders can improve outputs through iterative updates. Especially, convLSTM \cite{shi2015convolutional} and convGRU \cite{ballas2015delving} integrate convolution operations with RNNs to effectively learn spatial features within data and enhance computational efficiency, making them well-suited for vision tasks. Due to these advantages, various researchers have utilized convLSTM and convGRU as recurrent decoders for different tasks. PredRNN \cite{wang2017predrnn} and PredRNN++ \cite{wang2018predrnn++} have utilized convLSTM to predict video frames. RAFT \cite{teed2020raft} has applied convGRU to an optical flow estimation model, and subsequently, many models have utilized an iterative estimator in the decoder \cite{zhou2023mvflow}. Inspired by these applications, we uniquely reconfigure convGRU to suit SCD and incorporate convGRU into the decoder of our model.

\section{Methodology}
\subsection{Foundation Model-based Encoder}

Fig. \ref{fig:main_fig} illustrates the architecture of TERDNet. The encoder extracts features from the intermediate layers of the model. This involves dividing the attention layers into four sections and extracting features from the last layer of each section. For instance, in a ViT-b model with 12 attention layers, features are extracted from layers 2, 5, 8, and 11.
Each selected block corresponds to one quarter of the encoder depth to provide representations from low- to high-level abstraction with minimal redundancy.
In contrast to CNN-based feature pyramids, where feature maps vary in resolution, transformer-based encoders produce features of uniform size. This uniformity enhances the consistency of feature representation. By preserving a consistent feature size, the encoder also enables seamless integration with subsequent modules and ensures that the extracted features retain rich and detailed information across all scales.

\subsection{Feature Fusion Module}

A critical aspect of SCD is understanding both the individual characteristics of each input image and the relationship between the pair. To achieve this, TERDNet employs a feature fusion module that combines correlation volumes with the feature maps of each image. 

For relationship information, we employ the correlation volume from the optical flow research \cite{dosovitskiy2015flownet}. Given feature maps extracted from images at times $t0$ and $t1$, denoted as $m_0$ and $m_1$ respectively, for a pixel $(x_1, x_2)$ in $m_0$, the correlation with $(2r+1)^2$ neighboring pixels around $(x_1, x_2)$ in $m_1$ is computed through dot products, representing the relationship between the two feature maps as follows:
\begin{align}
    c(x_1, x_2)=\sum_{o\in [-r, r]\times [-r, r]}\left\langle m_0(x_1+o), m_1(x_2+o)\right\rangle,
\end{align}
where $\left\langle \cdot, \cdot \right\rangle$ and $r$ denote the dot product and correlation range, respectively. While optical flow models require wide-range correlation calculations to detect fast and extensive object movements, SCD focuses on detecting the appearance, disappearance, or change of objects \cite{jst2015change,alcantarilla2018street}. In this case, a narrow-range correlation is sufficient.  Additionally, the correlation layer has a high computational demand, but limiting the search range enhances computational efficiency.

Next, to utilize the information from both images, we pass the feature map of each image through a convolution layer to reduce the number of channels. This reduction process ensures that the subsequent fusion steps are computationally efficient while retaining essential information. The reduced feature map is then fused with the correlation volume. We repeat this process for each layer of the feature pyramid, and the concatenated feature maps from all layers yield the final fused feature. This module preserves the distinct information present in each image as well as highlights the interactions between them, providing a comprehensive basis for detecting changes.

\subsection{Recurrent Decoder}

The advantage of the recurrent structure lies in its ability to iteratively update outputs, and it can reintegrate lost information at each iteration. Inspired by previous research \cite{teed2020raft} that applied the separable convGRU to the decoder in image comparison tasks, we reconfigure GRU for TERDNet and incorporate it into the decoder.

We propose a 3-gate-GRU that has three gates for SCD. We integrate the feature pyramid into the GRU, allowing the decoder to update outputs using information from all feature map layers. Our 3-gate-GRU adds a feature gate $p_t$ to the reset and update gates of the traditional GRU, which allows the model to incorporate input from the feature pyramid. The inclusion of the feature gate $p_t$ enables the model to weigh the importance of features from different layers dynamically. This capability is particularly valuable for SCD, as it allows the model to focus on the most relevant features. 
The Ratios for Reflection block computes a sigmoid-gated map $f_t$ by applying a learnable projection $P$ to the concatenated pyramid differences $(m_0^i - m_1^i)$, and $f_t$ serves as the pyramid-derived gating signal for the feature gate $p_t$.
Unlike hierarchical decoders using CNN backbones, which sequentially input features from different layers, our structure allows for prioritizing information of more critical layers from the feature pyramid. $p_t$ decides the extent to which $f_t$ influences the candidate state $\tilde{h_t}$, subsequently calculating the current state $h_t$ by considering the importance of the previous state $h_{t-1}$, current input $x_t$, and $f_t$ as follows:
\begin{align}
    f_t = \sigma \left( P * \mathrm{Concat}_i (m_0^i - m_1^i) \right), \\
    r_t = \sigma (W_r*x_t+U_r*h_{t-1}+F_r*f_t),  \\
    z_t = \sigma (W_z*x_t+U_z*h_{t-1}+F_z*f_t), \\
    p_t = \sigma (W_p*x_t+U_p*h_{t-1}+F_p*f_t), \\
    \tilde{h_t}=\tanh (W*x_t+U*(r_t \odot h_{t-1})+F*(p_t \odot f_t)), \\
    h_t = (1-z_t)h_{t-1}+z_t \tilde{h_t},
\end{align}
where $m_0^i$ denotes the $i$th layer of the feature pyramid $m_0$, $\mathrm{Concat}_i(\cdot)$ denotes channel-wise concatenation across pyramid levels, and $P$ is a learnable convolutional projection.

\subsection{Combined Upsampler}

Given the reduced resolution resulting from the transformer backbone encoder, TERDNet employs a combined upsampler to restore the original image resolution. Unlike traditional upsampling methods, which may struggle with the uniform output size of transformer-based features, our combined upsampler utilizes a sequence of convolution and bilinear interpolation steps. Our combined upsampler stacks blocks of $3\times 3$ convolution layers and bilinear interpolation to recover resolution. The output of the GRU decoder, initially $\frac{H}{16}\times \frac{W}{16} \times 512$, is upscaled to $H\times W\times 2$ resolution using four blocks, with convolution reducing channel numbers to 256, 128, 64, and finally 2. Each bilinear interpolation follows the convolution layers and gradually increases the resolution rather than expanding sixteenfold at once. This design improves upsampling efficiency and preserves the quality of the output change mask. This approach addresses the challenges posed by the uniform output size of transformer-based features and helps produce accurate and detailed change masks.

\subsection{Training Loss}
We train TERDNet with a sequential weighted cross-entropy loss, tailored to suit the recurrent decoder and SCD datasets. The integration of sequential and weighted cross-entropy losses ensures that the model is trained to handle the unique challenges of SCD. The sequential loss calculates the loss for all predictions, and later iterations are weighted more heavily in the total loss through the utilization of the weight factor $\gamma$ ($\gamma = 0.8$ in all experiments). On the other hand, pixels with changes tend to be less frequent than those without changes in SCD. To mitigate this class imbalance, we integrate the sequential loss for the 3-gate-GRU decoder with the weighted cross-entropy loss for SCD; the training loss for our network becomes as follows:
\begin{align}
    L = \sum_{k=1}^M \gamma^{M-k}\sum_{i=0}^N w_iy_i\log p_i^k,  \\
    w_i = {\sum_{j=0}^N n_j-n_i \over N\sum_{j=0}^N n_j},
\end{align}
where $p_i^k$ and $y_i$ represent the predicted probability of $k$th iteration and ground truth for the two classes, respectively, $w_i$ is the balance weight, and since the SCD task only distinguishes between the presence and absence of change, $N=1$. This comprehensive approach for the loss function configuration is critical for achieving high performance in SCD tasks, where the precision of change detection is paramount.

\section{Experiments}
\begin{table}[!t]
\centering
\begin{small}
\caption{\textbf{Comparison study results.} (a) F1-scores (\%), (b) mIoU (\%) of previous methods and the proposed model. S and C represent the IoU for the static and Change classes, respectively.}
\subtable[F1-scores on VL-CMU-CD and TSUNAMI]{
\label{tb:f1}
\begin{tabular}{lcc}
\toprule
\textbf{Method} & VL-CMU-CD & TSUNAMI  \\
\midrule
FC-EF & 44.6 & 77.7 \\
FC-Siam-diff & 65.3 & 79.5 \\
FC-Siam-conc & 65.6 & 81.6 \\
CSCDNet & 76.6 & 84.8 \\
DR-TANet & 75.1 & 84.5 \\
C-3PO & \underline{80.2} & 86.5 \\
ZSSCD & 51.6 & 56.5 \\
RobustSCD & 79.5 & \underline{88.1}\\
GeSCD & 75.4 & 72.8 \\
\midrule
Ours & \textbf{83.4} & \textbf{89.5} \\
\bottomrule
\end{tabular}}
\subtable[mIoU on PSCD and ChangeSim]{
\label{tb:miou}
\begin{tabular}{lcccccc}
\toprule
\multirow{2}[1]{*}{\textbf{Method}} & \multicolumn{3}{c}{PSCD} &  \multicolumn{3}{c}{ChangeSim} \\
\cmidrule(lr){2-4}\cmidrule(lr){5-7}
& S & C & mIoU & S & C & mIoU \\
\midrule
FC-EF & 83.7 & 48.8 & 66.3 & 70.2 & 20.3 & 45.2 \\
FC-Siam-diff & 88.9 & 55.7 & 72.3 & 80.0 & 25.9 & 53.0 \\
FC-Siam-conc & 86.3 & 57.2 & 71.8 & 80.1 & 26.1 & 53.1 \\
CSCDNet & 88.8 & 61.7 & 75.3 & 87.3 & 22.9  & 55.1 \\
DR-TANet & 89.4 & 60.3 & 74.9 & 89.0 & 24.3 & 56.7 \\
C-3PO & \underline{90.8} & \underline{67.2} & \underline{79.0} & \underline{90.4} & 28.8 & 59.6 \\
ZSSCD & 10.8 & 18.9 & 14.8 & 89.8 & 1.4 & 45.6 \\
RobustSCD & 64.5 & 47.8 & 56.2 & 71.8 & 22.9 & 47.4 \\
GeSCD & 85.1 & 39.2 & 62.1 & 89.8 & \underline{33.2} & \underline{61.5} \\
\midrule
Ours & \textbf{91.5} & \textbf{69.6} & \textbf{80.5} & \textbf{91.2} & \textbf{38.3} & \textbf{64.7} \\
\bottomrule
\end{tabular}}
\end{small}
\end{table}

\subsection{Settings}
\subsubsection{Datasets} 
For a comparative study between conventional SCD models and TERDNet, we utilize four datasets: VL-CMU-CD \cite{alcantarilla2018street}, TSUNAMI \cite{jst2015change}, PSCD \cite{sakurada2020weakly}, and ChangeSim \cite{park2021changesim}. 
We select datasets that are frequently used as benchmarks in SCD studies and cover diverse environments such as indoor and outdoor scenes. 
We train and evaluate a separate model for each benchmark using its official split. 
We follow each dataset’s official train/test split and use the dataset-provided input order $(t0,t1)$, since the ground-truth change masks are defined with respect to the image at $t1$.

\subsubsection{Baselines}
To ensure fairness and reproducibility, we compare TERDNet against 9 previous SCD models with publicly released code, including FC-EF \cite{daudt2018fully}, FC-Siam-diff \cite{daudt2018fully}, FC-Siam-conc \cite{daudt2018fully}, CSCDNet \cite{sakurada2020weakly}, DR-TANet \cite{chen2021dr}, C-3PO \cite{wang2023reduce}, ZSSCD \cite{cho2025zero}, RobustSCD \cite{lin2025robust}, and GeSCD \cite{kim2025towards}. 

\begin{figure}
\centering
\subfigure[Comparison study results on VL-CMU-CD]{
\includegraphics[width=\linewidth]{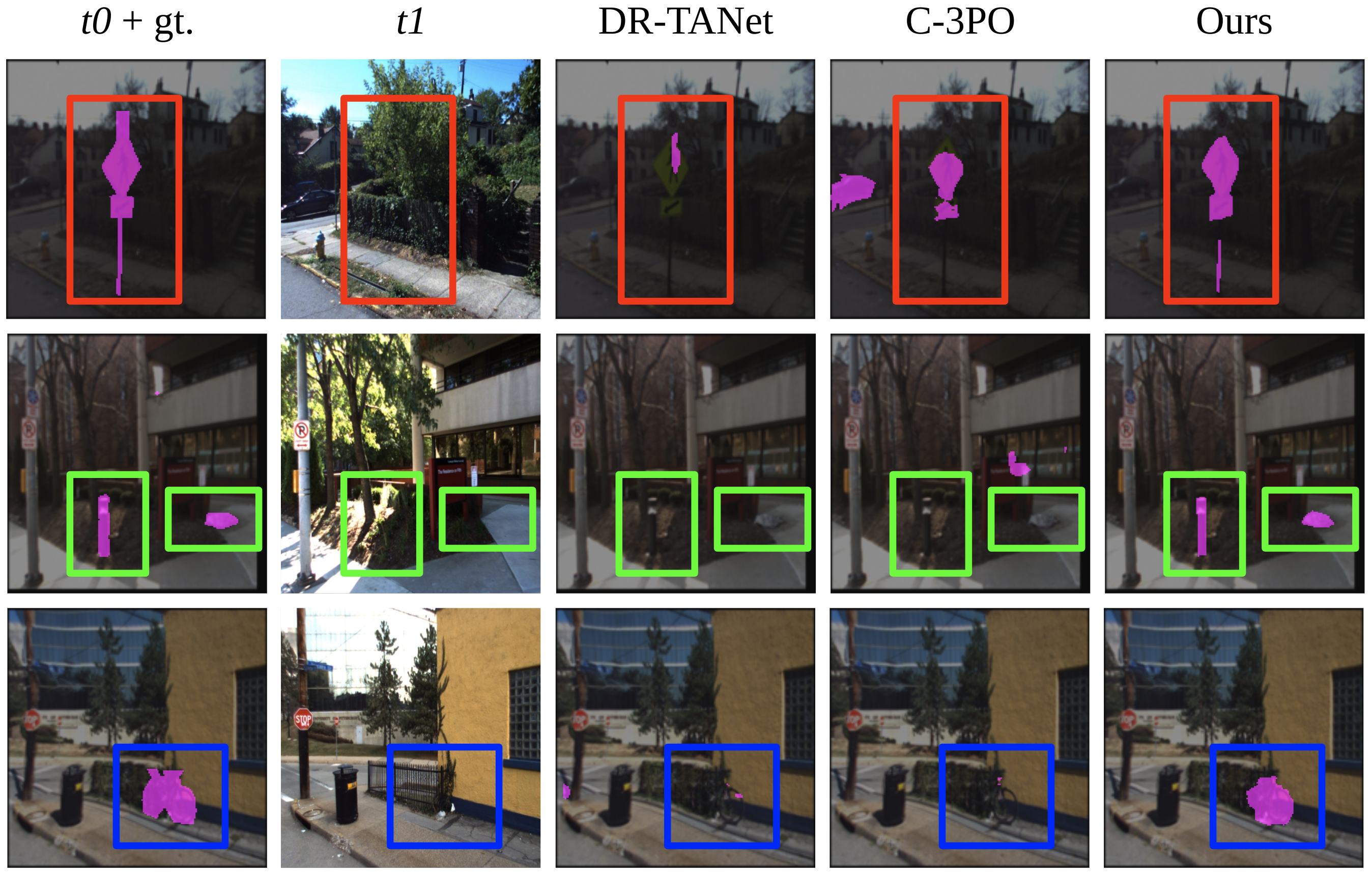}
}
\subfigure[Comparison study results on ChangeSim]{
\includegraphics[width=\linewidth]{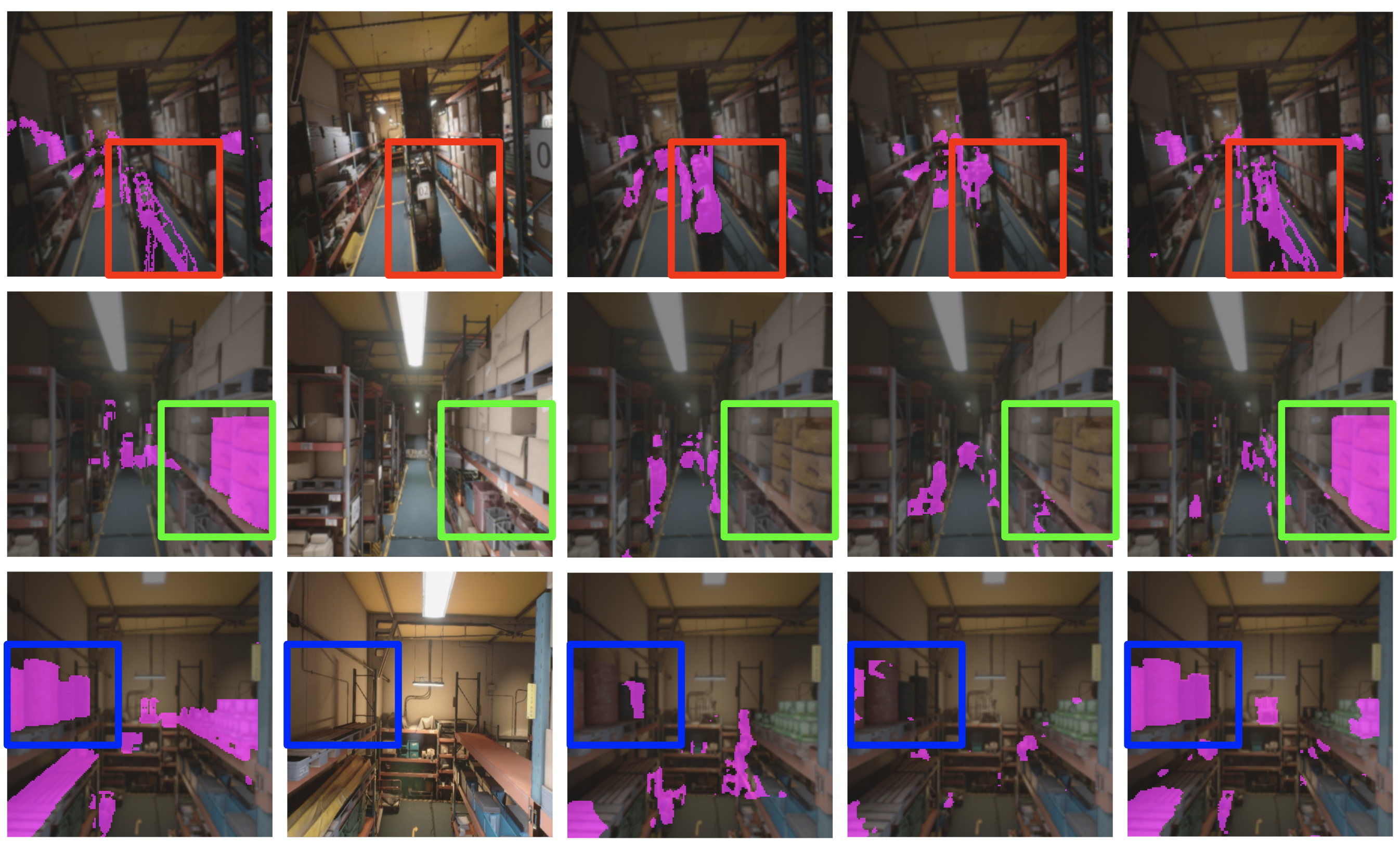}
}
\caption{
\textbf{Qualitative results of Comparative study on VL-CMU-CD, and ChangeSim. }\textcolor{red}{\textbf{Red}} boxes highlight thin or low-contrast structures,
\textcolor{green}{\textbf{green}} boxes precise localization of changes,
and \textcolor{blue}{\textbf{blue}} boxes region completion. Predicted masks from prior methods \cite{chen2021dr,wang2023reduce} and TERDNet visually indicate cleaner boundaries and more complete regions.
}
\label{fig:qualitative}
\end{figure}

\subsection{Comparative Study}
\label{sec:experiments}
\subsubsection{VL-CMU-CD and TSUNAMI}
We evaluated TERDNet using the F1-score metric on the VL-CMU-CD and TSUNAMI datasets. As shown in Table \ref{tb:f1}, TERDNet consistently outperformed prior approaches on both benchmarks. These improvements indicate that the proposed feature fusion and recurrent refinement allow the model to capture both everyday scene variations in VL-CMU-CD and the more abrupt changes present in TSUNAMI.

\subsubsection{PSCD and ChangeSim}
Since prior studies \cite{sakurada2020weakly, wang2023reduce} on PSCD and ChangeSim reported results in terms of mIoU, we adopt the same metric for comparison. As shown in Table \ref{tb:miou}, TERDNet achieved a higher mIoU than all previous approaches on PSCD. This result confirms its ability to handle real-world outdoor imagery with diverse change scenarios. On ChangeSim, a synthetic indoor benchmark, TERDNet also surpassed the baselines. Overall, these results suggest that the proposed design remains effective across different domains, such as real-world outdoor and synthetic indoor environments.

\subsubsection{Qualitative Comparison}
Fig. \ref{fig:qualitative} depicts qualitative comparison results. The qualitative results show that TERDNet can extract more sophisticated change masks. The conventional models often fail to generate complex masks (red boxes), accurately identify the locations of changes (green boxes), or adequately fill in the regions (blue boxes). This is particularly observable in scenes with intricate changes or minimal object displacement, which is challenging for these models to discern accurately. In contrast, TERDNet is robust even in detecting subtle changes where colors are similar or objects are thin and hard to capture (red and green boxes), and it comprehensively covers the changed areas (blue boxes). In summary, TERDNet is more resilient to the SCD challenges, such as discrepancies in viewpoint, illumination, and object positioning between the $t0$ and $t1$ images.
\begin{table}[!t]
\centering
\caption{\textbf{Ablation study results.} (a) The effect of the backbone encoder, (b) the effect of the pretraining method, (c) the effect of the finetuning method, (d) the effect of used feature types, (e) the effect of the decoder structure, and (f) the effect of the number of iterations.}
\subtable[The backbone encoder]{
\label{tb:encoder}
\begin{tabular}{lccc}
\toprule
\textbf{Architecture} & \textbf{Model} & \textbf{Parameters} & F1-score (\%) \\
\midrule
\multirow{2}[1]{*}{CNN} & ResNet18 & 
11M & 79.6 \\
 & ResNet50 & 26M & 80.9 \\
\midrule
\multirow{3}[1]{*}{Transformer} & ViT-b & 86M & 83.4 \\
 & ViT-l & 307M & \underline{83.7} \\
 & ViT-h & 632M & \textbf{84.0} \\
\bottomrule
\end{tabular}}
\subtable[The pretraining method]{
\label{tb:pretrain}
\begin{tabular}{lccc}
\toprule
\textbf{Method} & \textbf{Dataset} & Resolution & F1-score (\%) \\
\midrule
\multirow{2}[1]{*}{Supervised} & ImageNet & $384 \times 384$ & 60.0 \\
 & SA-1B & $1024 \times 1024$ & \textbf{83.4} \\
\midrule
\multirow{2}[1]{*}{MAE} & ImageNet & $224 \times 224$ & 57.2 \\
 & COCO & $224 \times 224$ & 56.0 \\
\midrule
\multirow{2}[1]{*}{DINOv2} & ImageNet & $784 \times 784$ & \underline{74.3} \\
 & COCO & $784 \times 784$ & \underline{74.3} \\
\bottomrule
\end{tabular}}
\subtable[The finetuning method]{
\label{tb:finetuning}
\begin{tabular}{lcc}
\toprule
\textbf{Method \qquad} & F1-score (\%) \\
\midrule
LLRD & 77.3 \\
LoRA & \underline{82.9} \\
VPT & 82.3 \\
\midrule
Frozen & \textbf{83.4} \\
\bottomrule
\end{tabular}}
\subtable[The used feature types]{
\label{tb:feature}
\begin{tabular}{lcc}
\toprule
\textbf{Method} & F1-score (\%) \\
\midrule
Feature Maps & \underline{82.3} \\
Local Corr & 81.2 \\
Global Corr & 81.3 \\
\midrule
Ours & \textbf{83.4} \\
\bottomrule
\end{tabular}}
\subtable[The decoder structure]{
\label{tb:decoder}
\begin{tabular}{lc}
\toprule
\textbf{Method} & F1-score (\%) \\
\midrule
Without GRU & 79.5 \\
With basic GRU & \underline{81.9} \\
\midrule
Ours & \textbf{83.4} \\
\bottomrule
\end{tabular}}
\subtable[The number of iterations]{
\label{tb:iterations}
\begin{tabular}{lc}
\toprule
\textbf{Iterations} & F1-score (\%) \\
\midrule
3 & 83.0 \\
5 & \textbf{83.4} \\
7 & \underline{83.2} \\
10 & 82.9 \\
\bottomrule
\end{tabular}}
\end{table}

\subsection{Ablation Study}
Through a series of experiments, we explore the effects of various pretraining and finetuning methods, the integration of a feature fusion module, the implementation of a recurrent structure, and the number of iterations in the decoder. In our ablation study, we employ the VL-CMU-CD dataset to examine the performance of each component of TERDNet.

\subsubsection{Backbone Encoder}
Table \ref{tb:encoder} compares the performance across different encoders. With a ResNet backbone, TERDNet already surpasses prior SCD models and confirms that the overall architecture is effective even without a foundation model encoder. Transformer backbones further improve performance, although larger variants give diminishing returns relative to their parameter count. Based on this trade-off, the ViT-b variant provides a practical balance between accuracy and efficiency.

\subsubsection{Pretraining Methods}
To investigate the effect of pretraining methods, we utilized supervised learning \cite{kirillov2023segment}, MAE \cite{he2022masked}, and DINOv2 \cite{oquab2023dinov2}, each trained on classification and segmentation tasks. In Table \ref{tb:pretrain}, different input resolutions were also tested, and models pretrained with sizes below $512 \times 512$ showed degraded performance. Increasing the resolution to $784 \times 784$ improved results, but models pretrained on the large-scale segmentation dataset SA-1B \cite{kirillov2023segment} achieved the best scores. 
Since dataset scale, supervision type, and input resolution are inherently coupled in publicly available checkpoints, this study highlights consistent trends across representative pretraining strategies rather than isolating a single factor.

\subsubsection{Finetuning Methods}
Table \ref{tb:finetuning} presents the performance results of various finetuning methods applied to the encoder. The best results were obtained with the frozen foundation model. This shows that the pretrained features already contain information essential for SCD. In contrast, full-parameter finetuning often caused overfitting and reduced performance. This issue was most evident with Layer-wise Learning Rate Decay (LLRD) \cite{clark2020electra}, which produced the lowest accuracy among the tested methods. Parameter-efficient approaches such as Low-Rank Adaptation (LoRA) \cite{hu2021lora} and Visual Prompt Tuning (VPT) \cite{jia2022visual} gave results close to the frozen baseline. Overall, in our experiments, keeping the encoder frozen was the most stable strategy for TERDNet.

\subsubsection{Feature Fusion Module}
Table \ref{tb:feature} shows the effect of the use of the feature map and correlation volume. The evaluation includes both local correlation volume, which calculates the correlation with nearby pixels, and global correlation volume, which calculates the correlation with all pixels. Our method, which utilized both feature maps and correlation volume, achieved the highest performance. These findings indicate that both the features of each image and the relationships between the two images are crucial in the SCD task, and TERDNet effectively leverages these aspects through its feature fusion module. Notably, there was little difference in performance between local and global correlation volumes, suggesting that in SCD, the relationship with distant pixels does not significantly impact the detection of changes since the task focuses solely on identifying the presence or absence of change.

\subsubsection{Recurrent Decoder}
Table \ref{tb:decoder} compares different decoder structures. Adding a standard GRU improved performance compared to using no recurrent structure, showing that recurrence helps refine the predictions. The 3-gate-GRU achieved even higher accuracy than the standard GRU. Unlike the standard GRU, which uses only reset and update gates, the 3-gate-GRU introduces an additional feature gate that controls how the fused features and the feature pyramid contribute to the update. This design produces more detailed change masks.

\subsubsection{Number of Iterations}
We examined the effect of the number of iterations in the 3-gate-GRU decoder. Table \ref{tb:iterations} shows that performance improved when the iteration count increased from 3 to 5. Beyond 5 iterations, however, the F1-score gradually declined, which suggests that the decoder began to overfit the training data. In our experiments, five iterations provided the best balance between accuracy and stability.

\begin{figure}
\begin{center}
\includegraphics[width=1\linewidth]{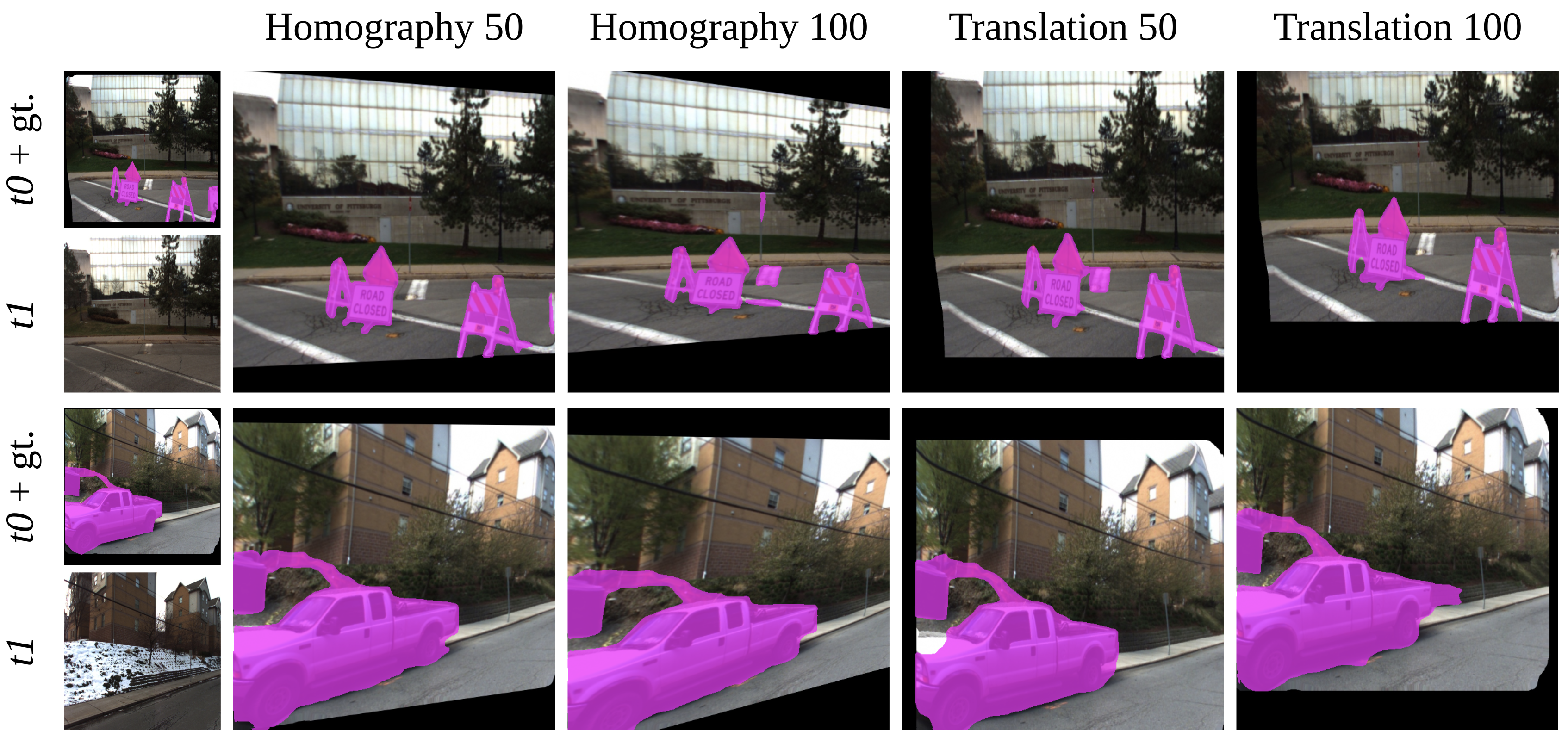}
\end{center}
\caption{
\textbf{Qualitative robustness evaluation under misalignment.} The $t0$ image is perturbed using homography or translation with magnitudes of 50 and 100 pixels. TERDNet generates consistent change masks without any task-specific finetuning, even when geometric distortions are introduced.
}
\label{fig:robustness}
\end{figure}

\subsection{Robustness to Misalignment}
We assess the robustness of TERDNet to misalignment by perturbing the $t0$ image through translation and homography, while keeping $t1$ unchanged. Fig. \ref{fig:robustness} presents qualitative results under perturbations of 50 and 100 pixels. Without additional finetuning, TERDNet generates consistent change masks and avoids false positives in static areas. These results suggest that the model can maintain robustness even under strong geometric distortions that are likely to appear in real deployments.

\begin{table}[!t]
\caption{\textbf{Comparison of SCD Models.}}
\label{tb:parameters}
\begin{small}
\resizebox{0.48\textwidth}{!}{
\begin{tabular}{lccccc}
\toprule
\textbf{Method} & Backbone Name & Backbone & Decoder & Total & GFLOPs \\
\midrule
FC-Siam-conc & U-Net & 0.4M & 0.6M & 1M & 43 \\
FC-Siam-diff & U-Net & 0.4M & 0.6M & 1M & 38 \\
\midrule
CSCDNet & ResNet18 & 11M & 83M & 94M & 337 \\
DR-TANet & ResNet18 & 11M & 22M & 33M & 54 \\
C-3PO & ResNet18 & 11M & 30M & 41M & 490 \\
\midrule
C-3PO & ResNet50 & 26M & 173M & 199M & 2715 \\
\midrule
C-3PO & VGG16 & 138M & 41M & 179M & 951 \\
\midrule
RobustSCD & ViT-s & 22M & 6M & 28M & 244 \\
\midrule
ZSSCD & ViT-h & 632M & 74M & 706M & 32928 \\
GeSCD & ViT-h & 632M & 9M & 641M & 56941 \\
\midrule
Ours (iters=3) & ViT-b & 86M & 16M & 102M & 1806\\
Ours (iters=5) & ViT-b & 86M & 16M & 102M & 2004\\
\bottomrule
\end{tabular}}
\end{small}
\end{table}
\vspace{-0.2cm}
\subsection{Efficiency Analysis of SCD Models}
\subsubsection{Model Parameters}
Table \ref{tb:parameters} compares the number of parameters across representative SCD models. U-Net \cite{daudt2018fully} and ResNet18-based models \cite{sakurada2020weakly, chen2021dr, wang2023reduce} have relatively small parameter counts, while VGG16-based approaches \cite{lei2020hierarchical,park2022dual,wang2023reduce} exceed 170M parameters. Recent transformer-based methods such as ZSSCD \cite{cho2025zero} and GeSCD \cite{kim2025towards} employ remarkably large ViT-h backbones, with over 600M parameters. TERDNet, in contrast, integrates a ViT-b backbone (86M) with a recurrent decoder (16M), totaling 102M parameters. This places it at a scale comparable to mid-sized ResNet-based models rather than the largest transformer-based approaches. The decoder remains compact while enabling iterative refinement through the 3-gate-GRU. Overall, TERDNet maintains a relatively modest parameter size compared to large ViT-h models, while achieving strong accuracy in the benchmark results reported in Section~\ref{sec:experiments}.

\subsubsection{Computational Complexity}
Table \ref{tb:parameters} also reports the computational complexity measured in GFLOPs. U-Net models require very few computations but typically underperform compared to larger networks. VGG16-based models \cite{lei2020hierarchical,park2022dual,wang2023reduce} require close to one thousand GFLOPs, while ResNet50-based C-3PO \cite{wang2023reduce} exceeds two thousand GFLOPs. In contrast, transformer models with ViT-h backbones, such as ZSSCD \cite{cho2025zero} and GeSCD \cite{kim2025towards}, require tens of thousands of GFLOPs and are therefore more computationally expensive. TERDNet requires 1806 GFLOPs with three iterations and 2004 GFLOPs with five iterations. While this is higher than U-Net or ResNet18-based models, it remains far below ViT-h approaches and is on the same order as ResNet50-based C-3PO. When considering both efficiency (Table \ref{tb:parameters}) and accuracy (Tables \ref{tb:f1}, \ref{tb:miou}), TERDNet offers a practical balance between computational cost and performance.

\section{Conclusion}
In this work, we presented TERDNet, a Transformer Encoder–Recurrent Decoder Network for scene change detection. TERDNet introduces a feature fusion module that combines correlation volumes with multi-level transformer features and emphasizes layer-wise importance, and a 3-gate-GRU decoder that iteratively refines predictions by integrating information across layers. Together with a combined upsampler, these components address the main limitations of existing SCD models, such as the lack of feature weighting and the reliance on single-step decoding. Through extensive experiments on four public benchmarks, TERDNet consistently outperformed prior approaches. The results show clear benefits from integrating transformer encoders with feature fusion and recurrent decoding, and the model achieves state-of-the-art performance. Beyond accuracy, evaluations on robustness demonstrated that TERDNet maintains stable performance under misalignment, which is essential for deployment in practical robotic and vision systems. Overall, this study highlights how carefully designed encoder fusion and recurrent decoding mechanisms can improve scene change detection. We believe the insights gained here can support future research on long-term perception in robotics and related vision tasks.

\small
\section*{ACKNOWLEDGMENT}
This research was partly supported by the National Research Foundation of Korea (NRF) grant funded by the Korea government (MSIT) (No. NRF-2022R1C1C1009989); by Institute of Information \& communications Technology Planning \& Evaluation (IITP) grant funded by the Korea government(MSIT) (No. RS-2022-II220926, Development of Self-directed Visual Intelligence Technology Based on Problem Hypothesis and Self-supervised Methods); by the National Research Council of Science \& Technology(NST) grant by the Korea government(MSIT) (No. GTL25041-000); and by the Korea Agency for Infrastructure Technology Advancement(KAIA) grant funded by the Ministry of Land Infrastructure and Transport (Grant RS-2023-00256888).

\bibliographystyle{IEEEtran}
\bibliography{ref}

\begin{thebibliography}{10}
\providecommand{\url}[1]{#1}
\csname url@rmstyle\endcsname
\providecommand{\newblock}{\relax}
\providecommand{\bibinfo}[2]{#2}
\providecommand\BIBentrySTDinterwordspacing{\spaceskip=0pt\relax}
\providecommand\BIBentryALTinterwordstretchfactor{4}
\providecommand\BIBentryALTinterwordspacing{\spaceskip=\fontdimen2\font plus
\BIBentryALTinterwordstretchfactor\fontdimen3\font minus
  \fontdimen4\font\relax}
\providecommand\BIBforeignlanguage[2]{{%
\expandafter\ifx\csname l@#1\endcsname\relax
\typeout{** WARNING: IEEEtran.bst: No hyphenation pattern has been}%
\typeout{** loaded for the language `#1'. Using the pattern for}%
\typeout{** the default language instead.}%
\else
\language=\csname l@#1\endcsname
\fi
#2}}

\bibitem{wang2023reduce}
G.-H. Wang, B.-B. Gao, and C.~Wang, ``How to reduce change detection to
  semantic segmentation,'' \emph{Pattern Recognition}, vol. 138, p. 109384,
  2023.

\bibitem{radke2005image}
R.~J. Radke, S.~Andra, O.~Al-Kofahi, and B.~Roysam, ``Image change detection
  algorithms: a systematic survey,'' \emph{IEEE Transactions on Image
  Processing}, vol.~14, pp. 294--307, 2005.

\bibitem{yew2021city}
Z.~J. Yew and G.~H. Lee, ``City-scale scene change detection using point
  clouds,'' in \emph{IEEE International Conference on Robotics and Automation},
  2021, pp. 13\,362--13\,369.

\bibitem{kannan2025zeroscd}
S.~S. Kannan and B.-C. Min, ``Zeroscd: Zero-shot street scene change
  detection,'' in \emph{IEEE International Conference on Robotics and
  Automation}, 2025, pp. 4665--4671.

\bibitem{rowell2024lista}
J.~Rowell, L.~Zhang, and M.~Fallon, ``Lista: Geometric object-based change
  detection in cluttered environments,'' in \emph{IEEE International Conference
  on Robotics and Automation}, 2024, pp. 3632--3638.

\bibitem{looper2023vsg}
S.~Looper, J.~Rodriguez-Puigvert, R.~Siegwart, C.~Cadena, and L.~Schmid, ``3d
  vsg: Long-term semantic scene change prediction through 3d variable scene
  graphs,'' in \emph{IEEE International Conference on Robotics and Automation},
  2023, pp. 8179--8186.

\bibitem{kirillov2023segment}
A.~Kirillov, E.~Mintun, N.~Ravi, H.~Mao, C.~Rolland, L.~Gustafson, T.~Xiao,
  S.~Whitehead, A.~C. Berg, W.-Y. Lo, \emph{et~al.}, ``Segment anything,'' in
  \emph{Proceedings of the IEEE/CVF International Conference on Computer
  Vision}, 2023, pp. 4015--4026.

\bibitem{wu2024lightweight}
Y.~Wu, F.~Paredes-Vall{\'e}s, and G.~C. De~Croon, ``Lightweight event-based
  optical flow estimation via iterative deblurring,'' in \emph{IEEE
  International Conference on Robotics and Automation}, 2024, pp.
  14\,708--14\,715.

\bibitem{gan2024rfl}
Y.~Gan, W.~Xuan, H.~Chen, J.~Liu, and B.~Du, ``Rfl-cdnet: Towards accurate
  change detection via richer feature learning,'' \emph{Pattern Recognition},
  vol. 153, p. 110515, 2024.

\bibitem{lin2017feature}
T.-Y. Lin, P.~Doll{\'a}r, R.~Girshick, K.~He, B.~Hariharan, and S.~Belongie,
  ``Feature pyramid networks for object detection,'' in \emph{Proceedings of
  the IEEE Conference on Computer Vision and Pattern Recognition}, 2017, pp.
  2117--2125.

\bibitem{cho2025zero}
K.~Cho, D.~Y. Kim, and E.~Kim, ``Zero-shot scene change detection,'' in
  \emph{Proceedings of the AAAI Conference on Artificial Intelligence},
  vol.~39, no.~3, 2025, pp. 2509--2517.

\bibitem{lin2025robust}
C.-J. Lin, S.~Garg, T.-J. Chin, and F.~Dayoub, ``Robust scene change detection
  using visual foundation models and cross-attention mechanisms,'' in
  \emph{IEEE International Conference on Robotics and Automation}, 2025, pp.
  8337--8343.

\bibitem{kim2025towards}
J.-W. Kim and U.-H. Kim, ``Towards generalizable scene change detection,'' in
  \emph{Proceedings of the Computer Vision and Pattern Recognition Conference},
  2025, pp. 24\,463--24\,473.

\bibitem{teed2020raft}
Z.~Teed and J.~Deng, ``Raft: Recurrent all-pairs field transforms for optical
  flow,'' in \emph{European Conference on Computer Vision}, 2020, pp. 402--419.

\bibitem{zhou2023mvflow}
S.~Zhou, X.~Jiang, W.~Tan, R.~He, and B.~Yan, ``Mvflow: Deep optical flow
  estimation of compressed videos with motion vector prior,'' in
  \emph{Proceedings of the ACM International Conference on Multimedia}, 2023,
  pp. 1964--1974.

\bibitem{sun2021nonlocal}
Y.~Sun, L.~Lei, X.~Li, H.~Sun, and G.~Kuang, ``Nonlocal patch similarity based
  heterogeneous remote sensing change detection,'' \emph{Pattern Recognition},
  vol. 109, p. 107598, 2021.

\bibitem{daudt2018urban}
R.~C. Daudt, B.~Le~Saux, A.~Boulch, and Y.~Gousseau, ``Urban change detection
  for multispectral earth observation using convolutional neural networks,'' in
  \emph{IEEE International Geoscience and Remote Sensing Symposium}, 2018, pp.
  2115--2118.

\bibitem{shi2015convolutional}
X.~Shi, Z.~Chen, H.~Wang, D.-Y. Yeung, W.-K. Wong, and W.-c. Woo,
  ``Convolutional lstm network: A machine learning approach for precipitation
  nowcasting,'' \emph{Advances in Neural Information Processing Systems},
  vol.~28, 2015.

\bibitem{Sun2020LUNetAL}
S.~Sun, L.~Mu, L.~Wang, and P.~Liu, ``L-unet: An lstm network for remote
  sensing image change detection,'' \emph{IEEE Geoscience and Remote Sensing
  Letters}, vol.~19, pp. 1--5, 2020.

\bibitem{huang2023background}
R.~Huang, R.~Wang, Q.~Guo, J.~Wei, Y.~Zhang, W.~Fan, and Y.~Liu,
  ``Background-mixed augmentation for weakly supervised change detection,'' in
  \emph{Proceedings of the AAAI Conference on Artificial Intelligence}, 2023,
  pp. 7919--7927.

\bibitem{gong2015change}
M.~Gong, J.~Zhao, J.~Liu, Q.~Miao, and L.~Jiao, ``Change detection in synthetic
  aperture radar images based on deep neural networks,'' \emph{IEEE
  Transactions on Neural Networks and Learning Systems}, vol.~27, pp. 125--138,
  2015.

\bibitem{daudt2018fully}
R.~C. Daudt, B.~Le~Saux, and A.~Boulch, ``Fully convolutional siamese networks
  for change detection,'' in \emph{IEEE International Conference on Image
  Processing}, 2018, pp. 4063--4067.

\bibitem{long2015fully}
J.~Long, E.~Shelhamer, and T.~Darrell, ``Fully convolutional networks for
  semantic segmentation,'' in \emph{Proceedings of the IEEE Conference on
  Computer Vision and Pattern Recognition}, 2015, pp. 3431--3440.

\bibitem{ronneberger2015u}
O.~Ronneberger, P.~Fischer, and T.~Brox, ``U-net: Convolutional networks for
  biomedical image segmentation,'' in \emph{Medical Image Computing and
  Computer-Assisted Intervention: International Conference}, 2015, pp.
  234--241.

\bibitem{sakurada2020weakly}
K.~Sakurada, M.~Shibuya, and W.~Wang, ``Weakly supervised silhouette-based
  semantic scene change detection,'' in \emph{IEEE International Conference on
  Robotics and Automation}, 2020, pp. 6861--6867.

\bibitem{chen2021dr}
S.~Chen, K.~Yang, and R.~Stiefelhagen, ``Dr-tanet: Dynamic receptive temporal
  attention network for street scene change detection,'' in \emph{IEEE
  Intelligent Vehicles Symposium}, 2021, pp. 502--509.

\bibitem{chen2017deeplab}
L.-C. Chen, G.~Papandreou, I.~Kokkinos, K.~Murphy, and A.~L. Yuille, ``Deeplab:
  Semantic image segmentation with deep convolutional nets, atrous convolution,
  and fully connected crfs,'' \emph{IEEE Transactions on Pattern Analysis and
  Machine Intelligence}, vol.~40, pp. 834--848, 2017.

\bibitem{oquab2023dinov2}
M.~Oquab, T.~Darcet, T.~Moutakanni, H.~Vo, M.~Szafraniec, V.~Khalidov,
  P.~Fernandez, D.~Haziza, F.~Massa, A.~El-Nouby, \emph{et~al.}, ``Dinov2:
  Learning robust visual features without supervision,'' \emph{arXiv preprint
  arXiv:2304.07193}, 2023.

\bibitem{kenton2019bert}
J.~D. M.-W.~C. Kenton and L.~K. Toutanova, ``Bert: Pre-training of deep
  bidirectional transformers for language understanding,'' in \emph{Proceedings
  of Human Language Technology: Conference of the North American Chapter of the
  Association of Computational Linguistics}, 2019, pp. 4171--4186.

\bibitem{vaswani2017attention}
A.~Vaswani, N.~Shazeer, N.~Parmar, J.~Uszkoreit, L.~Jones, A.~N. Gomez,
  {\L}.~Kaiser, and I.~Polosukhin, ``Attention is all you need,''
  \emph{Advances in Neural Information Processing Systems}, vol.~30, 2017.

\bibitem{dosovitskiy2020image}
A.~Dosovitskiy, L.~Beyer, A.~Kolesnikov, D.~Weissenborn, X.~Zhai,
  T.~Unterthiner, M.~Dehghani, M.~Minderer, G.~Heigold, S.~Gelly,
  \emph{et~al.}, ``An image is worth 16x16 words: Transformers for image
  recognition at scale,'' in \emph{International Conference on Learning
  Representations}, 2021.

\bibitem{mormille2023introducing}
L.~H. Mormille, C.~Broni-Bediako, and M.~Atsumi, ``Introducing inductive bias
  on vision transformers through gram matrix similarity based regularization,''
  \emph{Artificial Life and Robotics}, vol.~28, pp. 106--116, 2023.

\bibitem{liu2021swin}
Z.~Liu, Y.~Lin, Y.~Cao, H.~Hu, Y.~Wei, Z.~Zhang, S.~Lin, and B.~Guo, ``Swin
  transformer: Hierarchical vision transformer using shifted windows,'' in
  \emph{Proceedings of the IEEE/CVF International Conference on Computer
  Vision}, 2021, pp. 10\,012--10\,022.

\bibitem{carion2020end}
N.~Carion, F.~Massa, G.~Synnaeve, N.~Usunier, A.~Kirillov, and S.~Zagoruyko,
  ``End-to-end object detection with transformers,'' in \emph{European
  Conference on Computer Vision}, 2020, pp. 213--229.

\bibitem{ballas2015delving}
N.~Ballas, L.~Yao, C.~Pal, and A.~Courville, ``Delving deeper into
  convolutional networks for learning video representations,'' in
  \emph{International Conference on Learning Representations}, 2016.

\bibitem{wang2017predrnn}
Y.~Wang, M.~Long, J.~Wang, Z.~Gao, and P.~S. Yu, ``Predrnn: Recurrent neural
  networks for predictive learning using spatiotemporal lstms,'' \emph{Advances
  in Neural Information Processing Systems}, vol.~30, 2017.

\bibitem{wang2018predrnn++}
Y.~Wang, Z.~Gao, M.~Long, J.~Wang, and S.~Y. Philip, ``Predrnn++: Towards a
  resolution of the deep-in-time dilemma in spatiotemporal predictive
  learning,'' in \emph{International Conference on Machine Learning}, 2018, pp.
  5123--5132.

\bibitem{dosovitskiy2015flownet}
A.~Dosovitskiy, P.~Fischer, E.~Ilg, P.~Hausser, C.~Hazirbas, V.~Golkov, P.~Van
  Der~Smagt, D.~Cremers, and T.~Brox, ``Flownet: Learning optical flow with
  convolutional networks,'' in \emph{Proceedings of the IEEE International
  Conference on Computer Vision}, 2015, pp. 2758--2766.

\bibitem{jst2015change}
K.~Sakurada and T.~Okatani, ``Change detection from a street image pair using
  cnn features and superpixel segmentation,'' in \emph{British Machine Vision
  Conference}, 2015.

\bibitem{alcantarilla2018street}
P.~F. Alcantarilla, S.~Stent, G.~Ros, R.~Arroyo, and R.~Gherardi, ``Street-view
  change detection with deconvolutional networks,'' \emph{Autonomous Robots},
  vol.~42, pp. 1301--1322, 2018.

\bibitem{park2021changesim}
J.-M. Park, J.-H. Jang, S.-M. Yoo, S.-K. Lee, U.-H. Kim, and J.-H. Kim,
  ``Changesim: Towards end-to-end online scene change detection in industrial
  indoor environments,'' in \emph{IEEE/RSJ International Conference on
  Intelligent Robots and Systems}, 2021, pp. 8578--8585.

\bibitem{he2022masked}
K.~He, X.~Chen, S.~Xie, Y.~Li, P.~Doll{\'a}r, and R.~Girshick, ``Masked
  autoencoders are scalable vision learners,'' in \emph{Proceedings of the
  IEEE/CVF Conference on Computer Vision and Pattern Recognition}, 2022, pp.
  16\,000--16\,009.

\bibitem{clark2020electra}
K.~Clark, M.-T. Luong, Q.~V. Le, and C.~D. Manning, ``Electra: Pre-training
  text encoders as discriminators rather than generators,'' in
  \emph{International Conference on Learning Representations}, 2020.

\bibitem{hu2021lora}
E.~J. Hu, Y.~Shen, P.~Wallis, Z.~Allen-Zhu, Y.~Li, S.~Wang, L.~Wang, and
  W.~Chen, ``Lora: Low-rank adaptation of large language models,'' in
  \emph{International Conference on Learning Representations}, 2021.

\bibitem{jia2022visual}
M.~Jia, L.~Tang, B.-C. Chen, C.~Cardie, S.~Belongie, B.~Hariharan, and S.-N.
  Lim, ``Visual prompt tuning,'' in \emph{European Conference on Computer
  Vision}, 2022, pp. 709--727.

\bibitem{lei2020hierarchical}
Y.~Lei, D.~Peng, P.~Zhang, Q.~Ke, and H.~Li, ``Hierarchical paired channel
  fusion network for street scene change detection,'' \emph{IEEE Transactions
  on Image Processing}, vol.~30, pp. 55--67, 2020.

\bibitem{park2022dual}
J.-M. Park, U.-H. Kim, S.-H. Lee, and J.-H. Kim, ``Dual task learning by
  leveraging both dense correspondence and mis-correspondence for robust change
  detection with imperfect matches,'' in \emph{Proceedings of the IEEE/CVF
  Conference on Computer Vision and Pattern Recognition}, 2022, pp.
  13\,749--13\,759.

\end{thebibliography}

\end{document}